\documentclass[10pt,twocolumn,letterpaper]{article}

\usepackage{fancyhdr}
\usepackage{cuted}

\usepackage[pagenumbers]{cvpr} 

\usepackage[dvipsnames]{xcolor}

\definecolor{cvprblue}{rgb}{0.21,0.49,0.74}
\usepackage[pagebackref,breaklinks,colorlinks,citecolor=cvprblue]{hyperref}

\def\paperID{****} 
\def\confName{CVPR}
\def\confYear{2024}

\title{Fusion of regional and sparse attention in Vision Transformers}

\author{Nabil Ibtehaz$^{1,*}$ \quad Ning Yan$^{2}$ \quad Masood Mortazavi$^{2}$ \quad Daisuke Kihara$^{1}$ \vspace{0.3em} \\
{\normalsize $^1$Purdue University} \quad
{\normalsize $^2$Futurewei Technologies Inc.} \quad \\
{\normalsize \texttt{\{nibtehaz, dkihara\}@purdue.edu \{yan.ningyan, masood.mortazavi\}@futurewei.com} }
}

\begin{document}
\maketitle

\begin{abstract}

Modern vision transformers leverage visually inspired local interaction between pixels through attention computed within window or grid regions, in contrast to the global attention employed in the original ViT. Regional attention restricts pixel interactions within specific regions, while sparse attention disperses them across sparse grids. These differing approaches pose a challenge between maintaining hierarchical relationships vs.  capturing a global context. In this study, drawing inspiration from atrous convolution, we propose Atrous Attention, a blend of regional and sparse attention that dynamically integrates both local and global information while preserving hierarchical structures. Based on this, we introduce a versatile, hybrid vision transformer backbone called ACC-ViT, tailored for standard vision tasks. Our compact model achieves approximately 84\% accuracy on ImageNet-1K with fewer than 28.5 million parameters, outperforming the state-of-the-art MaxViT by 0.42\% while requiring 8.4\% fewer parameters.


\end{abstract}

\fancypagestyle{firstpage}
{
    \fancyhead[L]{Accepted as a Workshop Paper at T4V@CVPR2024}    
    \fancyhead[R]{}
}

\thispagestyle{firstpage}

\section{Introduction}

\let\thefootnote\relax\footnotetext{* Work done as an intern at Futurewei Technologies Inc.}

The early Vision Transformers (ViT) explored the feasibility of leveraging the seemingly infinite scalability \cite{Fedus2021SwitchSparsity} of text transformers for processing images but lacked sufficient inductive bias and were devoid of any vision-specific adjustments \cite{Touvron2020TrainingAttention}. Hence, they fell short to state-of-the-art CNN models \cite{Graham2021LeViT:Inference}. One particular aspect those former ViTs overlooked was local patterns, which usually carry strong contextual information \cite{Li2021LocalViT:Transformers}. This led to the development of windowed attention, proposed by Swin transformer \cite{Liu2021SwinWindows}, the first truly competent vision transformer model.This concept went through further innovations and eventually deviated towards two directions, namely, regional and sparse attention. In regional attention, windows of different sizes are considered to compute local attention at different scales \cite{Chu2021Twins:Transformers,Chen2021RegionViT:Transformers,Yang2021FocalTransformers}, whereas sparse attention computes a simplified global attention among pixels spread across a sparse grid \cite{Tu2022MaxViT:Transformer,Wang2021CrossFormer:Attention,Wang2023CrossFormer++:Attention}. Regional attention contains a sense of hierarchy \cite{Chen2021RegionViT:Transformers}, but the global view is harshly limited due to only considering a couple of regions to account for computational complexity \cite{Yang2021FocalTransformers}. On the contrary, sparse attention can compute a better approximate global context with reasonable computational cost, but sacrifices nested interaction across hierarchies \cite{Wang2021CrossFormer:Attention}. Although hierarchical information is immensely valuable \cite{Yang2022FocalNetworks}, owing to the access to a richer global context, sparse attention is used in state-of-the-art models and a relatively small-sized window attention can compensate for the limited local interaction \cite{Tu2022MaxViT:Transformer}.

\begin{figure}[h]
    \centering
    \includegraphics[width=\columnwidth]{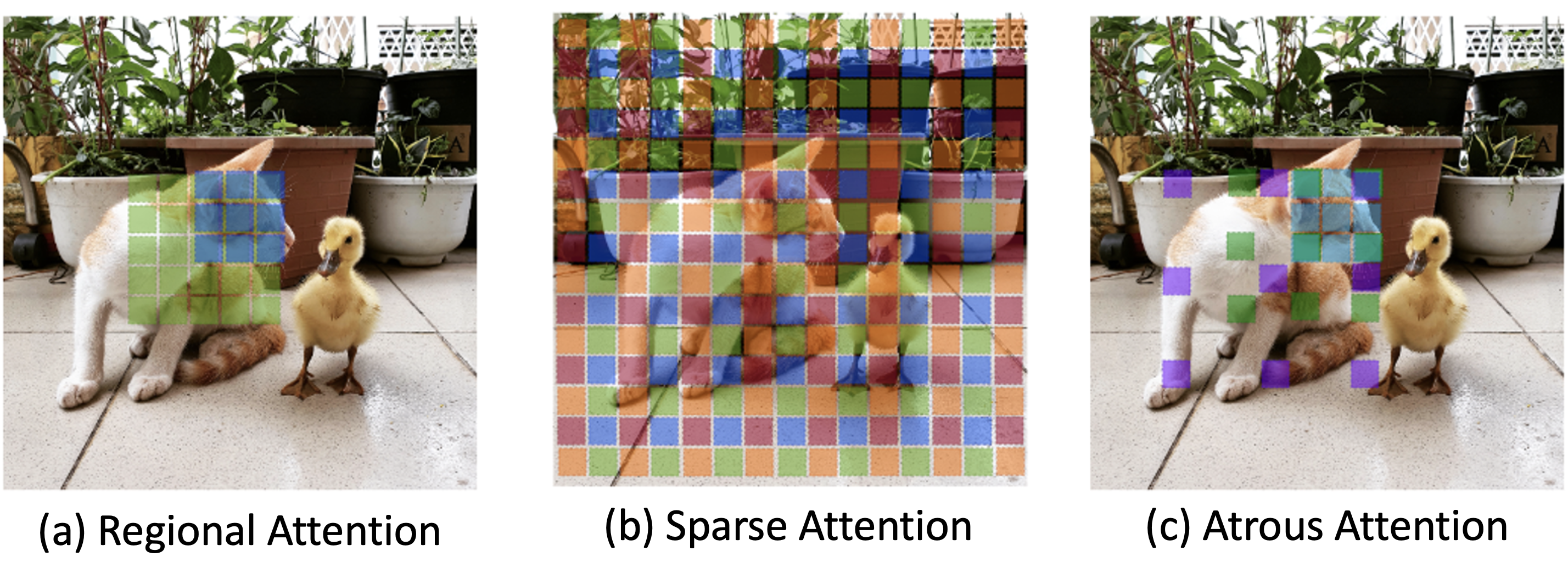}
    \caption{Simplified illustrations of different types of windowed attention mechanisms.}
    \label{fig:attn}
\end{figure}

In Fig. \ref{fig:attn}, we have presented simplified examples of regional and sparse attention. In regional attention (Fig. \ref{fig:attn}a), the cat is identified by two patches. The blue and the green patches inspect the head and most of the cat's body, respectively. By combining the information hierarchy, we can deduce that it is a cat. However, the legs and the tail have been missed by regional attention due to the high cost of computing attention over large regions. In the case of sparse attention (Fig. \ref{fig:attn}b), we can observe that the entire image is being analyzed by sets of grids marked by four different colors. Although seemingly, we can compute attention over the entire cat's body, there hardly exists information hierarchy and a good deal of irrelevant pixels are also considered. For instance, computing attention over the duck pixels probably does not contribute much to classifying the cat.

Based on the above intuitive analysis, we can make the following two observations. Firstly, it is preferable to cover more regions, with reasonable computational expense. Secondly, while global information is beneficial, we may probably benefit by limiting our receptive field up to a certain extent. Combining these two observations, we can thus induce sparsity in inspected regions or make the grids bounded. These reduce to dilated regions, which have been analyzed in atrous convolution \cite{Chen2016DeepLab:CRFs}. Therefore, taking inspiration from atrous convolution, we propose Atrous Attention. We can consider multiple windows, dilated at varying degrees, to compute attention and then combine them together. This solves the two issues concurrently, maintaining object hierarchy and capturing global context. For example, in Fig. \ref{fig:attn}c, it can be observed that 3 smaller patches with different levels of dilation are capable of covering all of the cat.

The concept of atrous or dilated convolution from traditional signal and image processing \cite{Holschneider1990ATransform} was properly introduced in computer vision by deeplab \cite{Chen2016DeepLab:CRFs}, which quickly gained popularity. By ignoring some consecutive neighboring pixels, atrous convolution increases the receptive field in deep networks \cite{Luo2017UnderstandingNetworks}. Moreover, Atrous spatial pyramid pooling can extract hierarchical features \cite{Chen2017RethinkingSegmentation}, which can also be employed adaptively based on conditions \cite{Qiao2020DetectoRS:Convolution}. However, in the ViT era, the use of atrous convolution has severely declined, limited only up to patch computation \cite{Huang2022AtrousInpainting}. Very recently the use of dilation in attention computation have been investigated. DiNat \cite{hassani2022dilated} involves computing attention maps with a fixed rate of dilation, whereas DilateFormer \cite{jiao2023dilateformer} leverages multiple dilation rates with smaller kernels.

We propose a novel attention mechanism for vision transformers, design a hybrid vision transformer based on that attention mechanism. We turn back to almost obscured atrous (or dilated) convolution, in the vision transformer era, and discover that the attributes of atrous convolution are quite beneficial for vision transformers. We thus design both our attention mechanism and convolution blocks based on the atrous convolution. In addition, taking inspiration from atrous spatial pyramid pooling (ASPP) \cite{Chen2016DeepLab:CRFs}, we experiment with parallel design, deviating from the existing style of stacking different types of transformer layers \cite{Tu2022MaxViT:Transformer}.

\section{Methods}
\label{sec:methods}

\begin{figure*}
    \centering
    \includegraphics[width=0.9\textwidth]{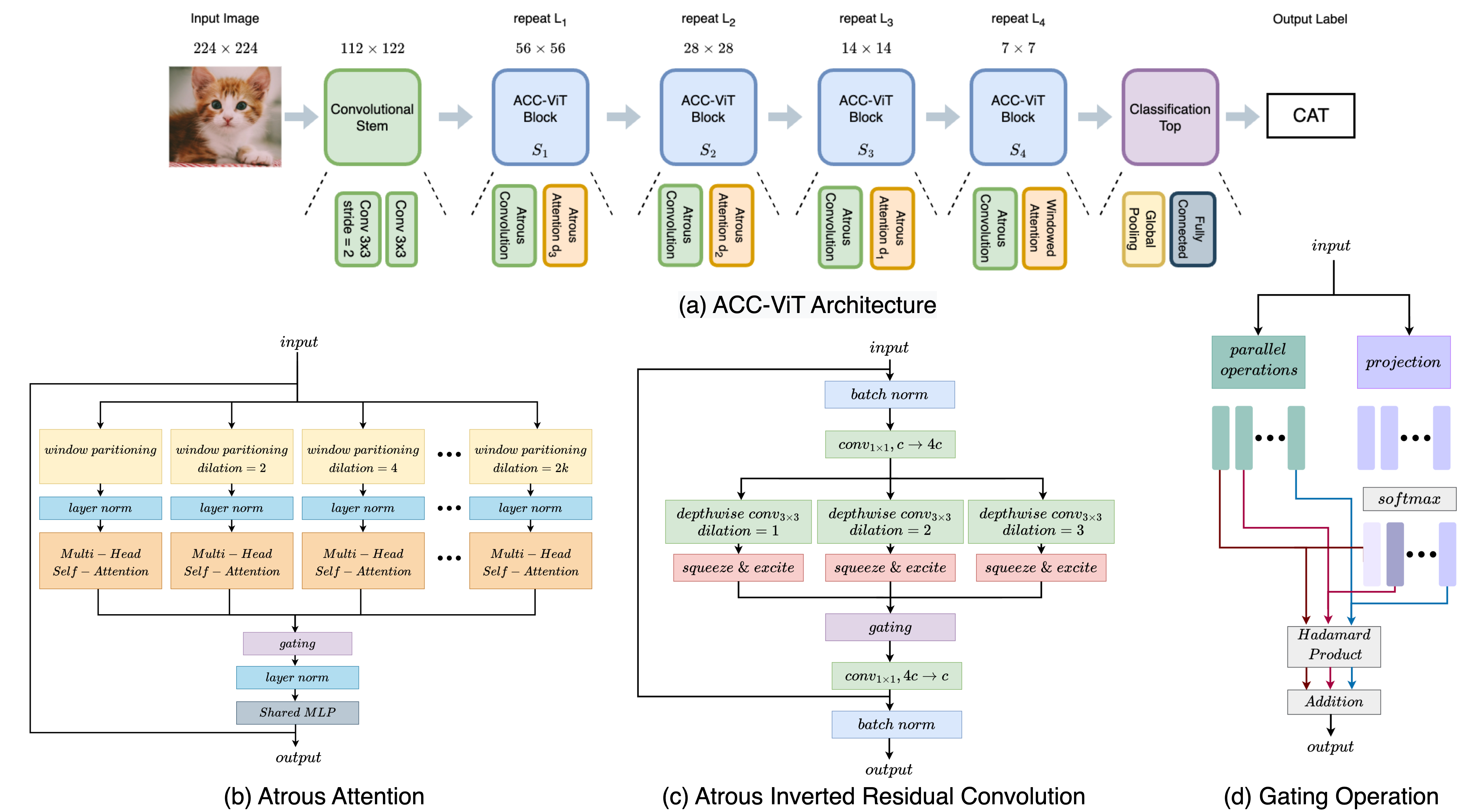}
    \caption{ACC-ViT model architecture and its components.}
    \label{fig:mdl-dia}
\end{figure*}

\subsection{Atrous Attention}

As a fusion between sparse and regional attention, we propose atrous attention (Fig. 1). We take motivation from atrous convolution \cite{Chen2016DeepLab:CRFs}, which drops some rows and columns from an image to effectively increase the receptive field, using similar number of parameters. This enables us to inspect over a global context with reasonable computational complexity, while also retaining the hierarchical information by considering windows at multiple dilation.

For an input featuremap $x \in \mathcal{R}^{c,h,w}$, where $c,h,w$ refers to the number of channels, height, and width, respectively, we compute windows at different levels of dilation as,
\begin{equation}
    x_{atr-k}[c,i,j] = x[c,i+2^k-1,j+2^k-1]
\end{equation}

Here, we only consider dilation rates as powers of 2, as they can be efficiently computed on GPU using einops operation \cite{Rogozhnikov2022Einops:Notation}. Similar to atrous convolution, all the values can be captured by shifting or sliding the dilated patterns. We apply windowed multi-head self-attention with relative positional embedding \cite{Shaw2018Self-AttentionRepresentations}, on each of the windows computed at different levels of dilation ($x_{atr-k}$) along with the undilated input ($x$). 



\subsection{Gating Operation for Adaptive Fusion of Different Branches}
\label{sec:gating}
Proper aggregation of hierarchical context can substantially improve the visual representation of vision models \cite{Yang2022FocalNetworks}. Therefore, we have developed a simple, lightweight, and adaptive gating operation to merge the features extracted from different levels of dilation.

Based on the input featuremap, $x$, $g \in [0 \sim 1]$ gating weight factors are computed for each output branch using a single linear layer coupled with softmax operation. We hypothesize that based on the input, the gating operation understands which portions of the different computed features should be emphasized or discarded.

For the different output braches $y_i$, corresponding weight factors $g_i$ are computed and the fused featuremap $y_{fused}$ can be computed from a Hadamard product.
\begin{equation}
    y_{fused} = \sum_{i=1}^{k} g_i \odot y_i
\end{equation}

\subsection{Shared MLP Layer across Parallel Attentions}

In our formulation, we apply a shared MLP layer on the fused attention map, unlike conventional transformer layers, where MLP layers are used after each attention operation. This simultaneously reduces the computational complexity, and makes the learning easier for the model (as seen in ablation). We conjecture that having a shared MLP reduces the degree of variation the model needs to reconcile.

\subsection{Parallel Atrous Inverted Residual Convolution}

Inverted residual convolution is the \textit{de facto} choice of convolution in vision transformers \cite{Tu2022MaxViT:Transformer,Yang2022MOAT:Models}. However, in order to exploit sparsity and hierarchy throughout the model, we replace the intermediate depthwise separable convolution with 3 parallel atrous, depthwise separable convolution, with dilation rates of 1, 2, and 3, respectively. The outputs of the three convolutional operations are merged through gating. Thus, formally our proposed Atrous Inverted Residual Convolution block can be presented as follows:
\begin{equation}
    y = x + SE(Conv_{1 \times 1}^{c\times 0.25}( G (DConv_{3 \times 3}^{dil=1,2,3} (Conv_{1 \times 1}^{c\times 4}(x)))))
\end{equation}
Here, $SE$ and $G$ refers to squeeze-and-excitation block \cite{Hu2018Squeeze-and-ExcitationNetworks} and gating operation, respectively. The subscripts of $Conv$ and $DConv$ denote the kernel size, and the superscripts refer to the expansion ratio or dilation rates, where applicable.

\begin{table*}[h]
\centering
\caption{Transfer learning experiment results on 3 medical image datasets.}

\label{tbl:res_med}
\begin{tabular}{c|cccc|cccc|cccc}
\cline{2-13}
                                & \multicolumn{4}{c|}{HAM10000} & \multicolumn{4}{c|}{EyePACS} & \multicolumn{4}{c}{BUSI} \\ \cline{2-13} 
                                & Pr   & Re   & F1  & Acc  & Pr   & Re   & F1   & Acc  & Pr   & Re   & F1  & Acc  \\ \hline
\multicolumn{1}{c|}{Swin-T}     &  82.34   &  75.93   & 78.37    &   88.32   &    \textbf{67.37}  &  48.94    &  52.85  & 83.05   &   75.18   &  75.70    &     75.07 &     77.56  \\
\multicolumn{1}{c|}{ConvNeXt-T} &   83.69   &   76.84   &   79.48  &  89.57    &  66.82    &  52.30    &  56.20    &   83.90   &    79.75  &     \textbf{79.65} &  \textbf{79.42}   &   \textbf{82.05}   \\
\multicolumn{1}{c|}{MaxVit-T}   &  64.45    &  57.87    &  59.71   &   84.62   &  64.63    &   51.43   & 53.99     &   83.67   &   72.30   &    63.75  &  66.46   &    73.08  \\
\multicolumn{1}{c|}{ACC-ViT-T}  &  \textbf{90.02}    &  \textbf{84.26}    &  \textbf{86.77}    &  \textbf{92.06}    &   66.42   &   \textbf{57.00}   &   \textbf{60.12}   &   \textbf{85.07}   &  \textbf{80.10}    &   76.30   &   77.80  &   80.77   \\ \hline
\end{tabular}
\end{table*}

\subsection{ACC-ViT}

Using the proposed attention and convolution blocks, we have designed ACC-ViT, a hybrid, hierarchical vision transformer architecture. Following conventions, the model comprises a convolution stem at the input level, 4 ACC-ViT blocks, and a classification top \cite{Tu2022MaxViT:Transformer}. The stem downsamples the input image, making the computation of self-attention feasible. The stem is followed by 4 ACC-ViT blocks, which are built by stacking Atrous Convolution and Attention layers, and the images are downsampled after each block. Since the image size gets reduced, we use 3,2,1 and 0 (no dilation) levels of Atrous Attention in the 4 blocks, respectively. Finally, the classifier top is based on global average pooling and fully connected layers, as used in contemporary architecture \cite{Tu2022MaxViT:Transformer,Liu2021SwinWindows,Liu2022A2020s}. A diagram of our proposed architecture is presented in Fig. \ref{fig:mdl-dia}.

Following the conventional practices, we have designed several variants of ACC-ViT, having different levels of complexity, the configurations are presented in the appendix.

\section{Results}
\label{sec:results}

\subsection{ImageNet-1k Image Classification}

\begin{figure}[h]
    \centering
\includegraphics[width=\columnwidth]{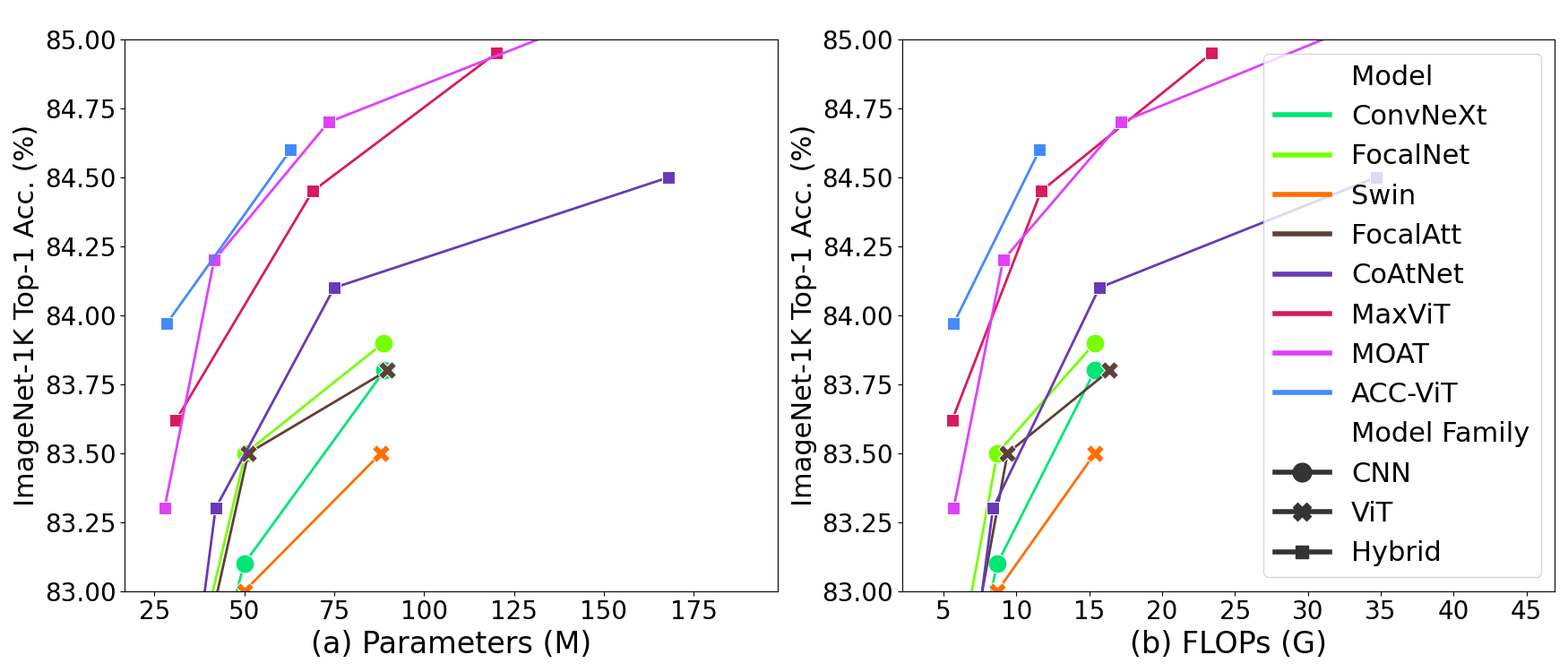}
    \caption{ACC-ViT performs competitively against state-of-the-art models on ImageNet-1K.}
    \label{fig:pareto}
\end{figure}

We have trained the different variants of ACC-ViT on the standard image classification task, using the ImageNet-1K dataset \cite{Deng2009ImageNet:Database}. The models were trained using a combination of the hyperparameter values recommended by Torchvision \cite{TorchVisionmaintainersandcontributors2016TorchVision:Library} and as adopted in recent works \cite{Yang2022FocalNetworks, Tu2022MaxViT:Transformer, Yang2022MOAT:Models}. Under the conventional supervised learning paradigm, the models were trained on $224 \times 224$ resolution images, without any extra data. The top-1 validation accuracy achieved by ACC-ViT is presented in Fig. \ref{fig:pareto}, along with a comparison against popular state-of-the-art models.

ACC-ViT outperforms the state-of-the-art MaxViT and MOAT models within similar ranges of parameters and FLOPs. In tiny and small models, using a similar amount of flops, ACC-ViT achieves $0.35\%$ and $0.15\%$ higher accuracy than MaxViT, despite having $9.15\%$ and $9.7\%$ less parameters. For all the other models, the performance of ACC-ViT is even more impressive. The tiny and small versions of ACC-ViT are more accurate than the small and base variants of the other models, respectively.

\subsection{Transfer Learning Experiment on Medical Image Datasets}

One of the most useful applications of general vision backbone models is transfer learning, where ViT models have demonstrated remarkable efficacy \cite{Zhou2021ConvNetsTransferable}. In order to evaluate the transfer learning capability of ACC-ViT, we selected medical image datasets for the diversity and associated challenges \cite{Panayides2020AIDirections}. 3 different medical image datasets of different modalities and size categories were selected, as a means to critically assess how transferable the visual representations are under different configurations. The images from HAM10000 \cite{Tschandl2018TheLesions} (skin melanoma, 10,015 images), EyePACS \cite{Cuadros2009EyePACS:Screening} (diabetic retinopathy, 5,000 images) and BUSI \cite{Al-Dhabyani2020DatasetImagesb} (breast ultrasound, 1,578 images) datasets were split into 60:20:20 train-validation-test splits randomly, in a stratified manner. The tiny versions of Swin, ConvNext, MaxViT and ACC-ViT, pretrained on ImageNet-1K were finetuned for 100 epochs and the macro precision, recall, F1 scores along with the accuracy on the test data were computed (Table \ref{tbl:res_med}). It can be observed that ACC-ViT outperformed the other models in most of the metrics, which was more noticeable on the larger datasets. For the small dataset, BUSI, ConvNeXt turned out to be the best model, ACC-ViT becoming the second best. This is expected as it has been shown in medical imaging studies that convolutional networks perform better than transformers on smaller datasets \cite{Ibtehaz2023ACC-UNet:Forthe2020s}. Out of all the models, MaxViT seemed to perform the worst, particularly for the rarer classes, which probably implies that the small-scale dataset was not sufficient to tune the model's parameters sufficiently, (please refer to the appendix for class-specific metrics).

\subsection{Model Interpretation}
\begin{figure}[h]
    \centering
    \includegraphics[width=\columnwidth]{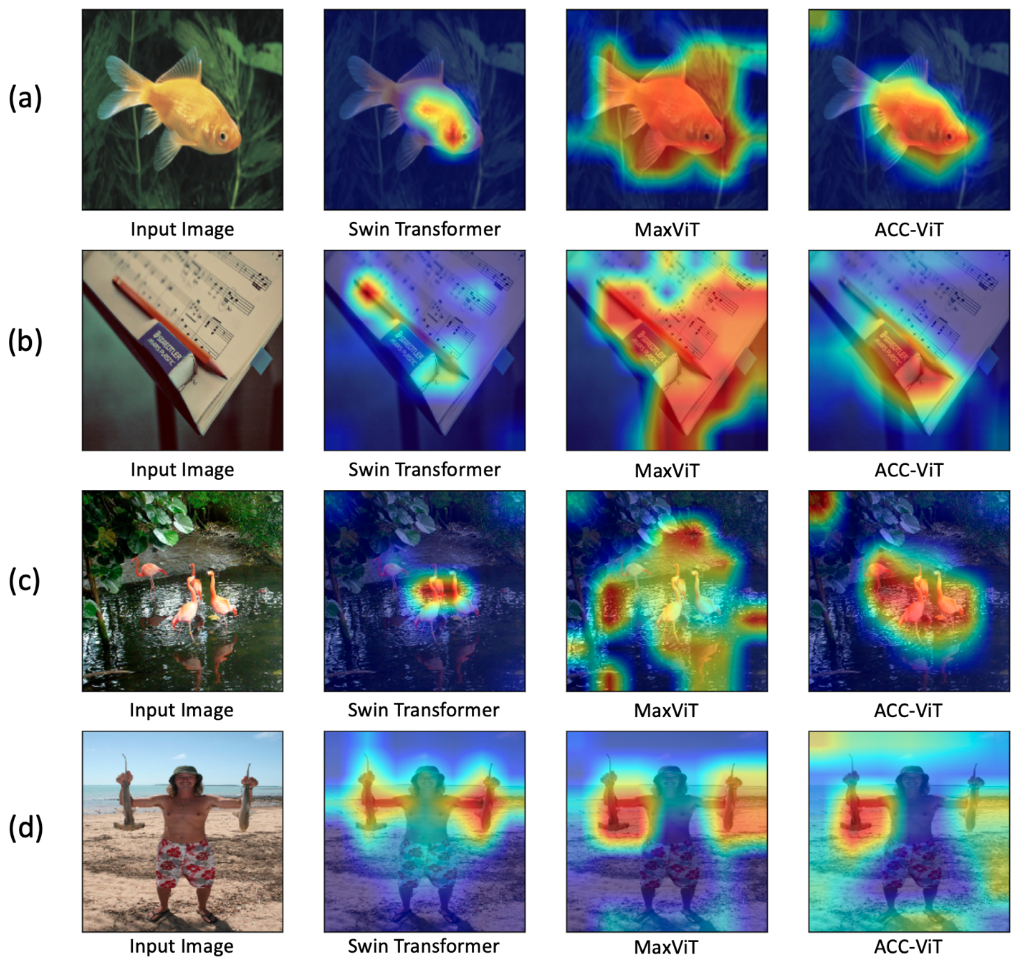}
    \caption{Model Interpretation using Grad-CAM.}
    \label{fig:grad-cam}
\end{figure}

In order to interpret and analyze what the model learns, we applied Grad-CAM \cite{Selvaraju2016Grad-CAM:Localization} on Swin, MaxViT and ACC-ViT (Fig. \ref{fig:grad-cam}). The class activation maps revealed that MaxViT tends to focus on irrelevant portions of the image, probably due to grid attention. On the contrary, Swin  seemed to often focus on a small local region. ACC-ViT apparently managed to inspect and distinguish the entire goldfish and eraser of (Fig. \ref{fig:grad-cam}a and b). Moreover, when classifying the flamingo from the example of Fig. \ref{fig:grad-cam}c, ACC-ViT focused on the entire flock of flamingos, whereas Swin and MaxViT focused on a subset and irrelevant pixels, respectively. Interestingly, when classifying  Fig. \ref{fig:grad-cam}d as hammerhead, ACC-ViT only put focus on the hammerhead fish, whereas the other transformers put emphasis on both the fishes.

\section{Conclusion}

In this work, we have developed a novel attention mechanism for vision transformers, by fusing the concepts of regional and sparse attention. Seeking inspiration from atrous convolution, we have designed a sparse regional attention mechanism, named Atrous Attention. Our proposed hybrid ViT architecture, ACC-ViT, maintains a balance between local and global, sparse and regional information throughout the model.

\clearpage
\clearpage
\clearpage

{
    \small
    \bibliographystyle{ieeenat_fullname}
    \bibliography{references}
}


\appendix

\clearpage
\maketitlesupplementary

\appendix

\section{Architectural Details}
\begin{strip}
\subsection{Efficient dilated window generation for Atrous Attention using einops}

Our proposed Atrous Attention depends on computing dilated windows efficiently. We have devised a compact einops operation \cite{Rogozhnikov2022Einops:Notation} to partition the images or featuremaps into dilated windows. 

$x-attr1$, $x-attr2$, and $x-attr3$, which are dilated versions of input $x$ with dilation rates of  2, 4, and 8, respectively, can be computed as:

\begin{verbatim}
    x-attr1 = rearrange(x, 'b c (h hs) (w ws) -> (b hs ws) c h w', hs=2, ws=2)
    x-attr2 = rearrange(x, 'b c (h hs) (w ws) -> (b hs ws) c h w', hs=4, ws=4)
    x-attr3 = rearrange(x, 'b c (h hs) (w ws) -> (b hs ws) c h w', hs=8, ws=8)
\end{verbatim}

In our ACC-ViT we consider dilation rates up to 8, therefore it suffices to compute up to $x-attr3$.

This operation effectively extracts $k^2$ dilated windows, partitioned at a dilation rate of $k$. The extracted windows thus can be used for computing windowed attention without any modification to the attention mechanism.

The original image or featuremap can be retrieved from the partitioned windows through a departition operation implemented using the inverse einops operation. 

\begin{verbatim}
    x = rearrange(x-attr1, '(b hs ws) c h w -> b c (h hs) (w ws)', hs=2, ws=2)
    x = rearrange(x-attr2, '(b hs ws) c h w -> b c (h hs) (w ws)', hs=4, ws=4)
    x = rearrange(x-attr3, '(b hs ws) c h w -> b c (h hs) (w ws)', hs=8, ws=8)
\end{verbatim}

\subsection{Overview of the different ACC-ViT layers}

The ACC-ViT layers are based on our proposed Atrous Attention, Atrous Inverted Residual Convolution, and Gating. The convolution operations are identical in all the stages. However. the different stages of ACC-ViT layers use different degrees of Atrous Attention, due to the change in resolution of featuremaps.

\begin{enumerate}[label=(S{{\arabic*}})]
    \item \textbf{Block 1} : Atrous Attention with dilation rates of 2, 4, and 8 are used, along with undilated windowed attention.
    \item \textbf{Block 2} : Atrous Attention with dilation rates of 2, and 4 are used, along with undilated windowed attention.
    \item \textbf{Block 3} : Atrous Attention with a dilation rate of 2 is used, along with undilated windowed attention.
    \item \textbf{Block 4} : Only undilated windowed attention is used.
\end{enumerate}
\end{strip}

\subsection{ACC-ViT configurations}

\begin{table}[!h]
\centering
\caption{Different Configurations of ACC-ViT model.}
\label{tbl:arch}
\begin{tabular}{l|c|ccc|ccc}
\hline
\textbf{Stage} & \textbf{Size} & \textbf{tiny} & \textbf{small} & \textbf{base} & \textbf{nano} & \textbf{pico} & \textbf{femto} \\ \hline
\textbf{S0: Stem}     & 1/2  & B=2, C=64  & B=2, C=64  & B=2, C=64   & B=2, C=64  & B=2, C=48  & B=2, C=32  \\
\textbf{S1: Block 1}  & 1/4  & B=2, C=64  & B=2, C=96  & B=4, C=96   & B=1, C=64  & B=1, C=48  & B=1, C=32  \\
\textbf{S2: Block 2}  & 1/8  & B=3, C=128 & B=3, C=192 & B=6, C=192  & B=2, C=128 & B=2, C=96  & B=2, C=64  \\
\textbf{S3: Block 3}  & 1/16 & B=6, C=256 & B=6, C=384 & B=14, C=384 & B=4, C=256 & B=4, C=192 & B=4, C=128 \\
\textbf{S4: Block 4}  & 1/32 & B=2, C=512 & B=2, C=768 & B=2, C=768  & B=1, C=512 & B=1, C=384 & B=1, C=256 \\
\textbf{\#params (M)} &      & 28.367     & 62.886     & 103.576     & 16.649     & 9.55       & 4.4        \\
\textbf{FLOPs (G)}    &      & 5.694      & 11.59      & 22.316      & 3.812      & 2.217      & 1.049     \\ \hline
\end{tabular}
\end{table}

\clearpage

\section{Implementation Details}

\begin{strip}

Recent state-of-the-art models make the implementation and trained-weights public. However, the training pipeline script is hardly accessible, which prevents the identical and consistent training of new models. As a result, there have been community efforts to retrain the models using standardized protocols, maintaining reproducibility. TorchVision \cite{TorchVisionmaintainersandcontributors2016TorchVision:Library} and timm \cite{Wightman2019PyTorchModels} are the two most notable open-source projects in this regard.

Our ACC-ViT implementation and ImageNet training pipeline are based on TorchVision v0.15.2 (\url{https://pypi.org/project/torchvision/0.15.2/}). Furthermore, we used the ImageNet-1K trained weights of the baseline models, released by TorchVision. The baseline model weights are summarized in Table \ref{tbl:trchvsn}.

\end{strip}

\begin{table}[]
\centering
\caption{TorchVision weights of baseline models}
\label{tbl:trchvsn}
\begin{tabular}{c|c|c|c|c}
\hline
Model &
  TorchVision weight &
  \begin{tabular}[c]{@{}c@{}}IN-1K Acc\\ (reproduced)\end{tabular} &
  \begin{tabular}[c]{@{}c@{}}IN-1K\\ (published)\end{tabular} &
  Link \\ \hline
Swin &
  Swin\_T\_Weights.IMAGENET1K\_V1 &
  81.474\% &
  81.3\% &
  \begin{tabular}[c]{@{}c@{}}https://download.pytorch.org/models/\\ swin\_t-704ceda3.pth\end{tabular} \\
ConvNeXt &
  ConvNeXt\_Tiny\_Weights.IMAGENET1K\_V1 &
  82.520\% &
  82.1\% &
  \begin{tabular}[c]{@{}c@{}}https://download.pytorch.org/models/\\ convnext\_tiny-983f1562.pth\end{tabular} \\
MaxViT &
  MaxVit\_T\_Weights.IMAGENET1K\_V1 &
  83.7\% &
  83.6\% &
  \begin{tabular}[c]{@{}c@{}}https://download.pytorch.org/models/\\ maxvit\_t-bc5ab103.pth\end{tabular} \\ \hline
\end{tabular}
\end{table}

\section{Experimental Details}

\subsection{ImageNet-1K Training}


The ImageNet-1K dataset \cite{Deng2009ImageNet:Database} is a benchmark dataset containing images from 1000 classes. It provides 1.2 million and 50,000 images for training and validation, respectively. We trained the ACC-ViT models mostly based on the TorchVision defaults and the recipe for MaxViT \url{https://github.com/pytorch/vision/tree/main/references/classification#maxvit} and trained the models for 400 epochs with cosine annealing learning rate scheduler and exponential moving average (EMA). The batch size and learning rate were adjusted based on the computational limitations set by GPU memory. Taking inspiration from FocalNet \cite{Yang2022FocalNetworks} we considered both cutmix \cite{Yun2019CutMix:Features} and mixup \cite{Zhang2017Mixup:Minimization} augmentations.

The training hyperparameters are summarized in table \ref{tbl:hypimg}.

\begin{table}[h]
\centering
\caption{Hyperparameters for ImageNet training}
\label{tbl:hypimg}
\begin{tabular}{c|ccccc}
\hline
Hyperparameter        & femto    & pico    & nano    & tiny    & small    \\ \hline
image resolution      & \multicolumn{5}{c}{$224 \times 224$} \\
batch size            & 3072     & 2560    & 2048    & 1024    & 768      \\
stochastic depth      & 0.0      & 0.0     & 0.1     & 0.2     & 0.3      \\
learning rate         & 1e-3     & 1e-3    & 1e-3    & 1e-3    & 5e-4     \\
min learning rate     & 1e-5     & 1e-5    & 1e-5    & 1e-5    & 5e-6     \\
training epochs       & \multicolumn{5}{c}{400}                           \\
warm-up epochs       & \multicolumn{5}{c}{32}                           \\
Optimizer             & \multicolumn{5}{c}{adamw}                         \\
schedular             & \multicolumn{5}{c}{cosine annealing}              \\
loss function         & \multicolumn{5}{c}{softmax}                       \\
randaugment policy    & \multicolumn{5}{c}{ta-wide}                       \\
randaugment magnitude & \multicolumn{5}{c}{15}                            \\
randaugment sampler   & \multicolumn{5}{c}{3}                             \\
cutmix-alpha          & \multicolumn{5}{c}{0.8}                           \\
mixup-alpha           & \multicolumn{5}{c}{0.8}                           \\
label-smoothing       & \multicolumn{5}{c}{0.1}                           \\
gradient clip         & \multicolumn{5}{c}{1.0}                           \\
ema steps             & \multicolumn{5}{c}{32}                          \\ \hline 
\end{tabular}
\end{table}

\clearpage

\begin{strip}

The detailed training commands of the ACC-ViT models, based on TorchVision are as follows:

\begin{itemize}

\item ACC-ViT femto

\begin{verbatim}
CUBLAS_WORKSPACE_CONFIG=:16:8 torchrun --nproc-per-node=16 --nnodes=1 train.py 
--model accvit_f --epochs 400 --batch-size 192 --opt adamw --lr 1e-3 
--weight-decay 0.05 --lr-scheduler cosineannealinglr --lr-min 1e-5 
--lr-warmup-method linear --lr-warmup-epochs 32 --label-smoothing 0.1 
--mixup-alpha 0.8 --cutmix-alpha 0.8 --clip-grad-norm 1.0 --interpolation bicubic 
--auto-augment ta_wide --ra-magnitude 15 --ra-sampler --model-ema 
--val-resize-size 224 --val-crop-size 224 --train-crop-size 224 
--model-ema-steps 32 --transformer-embedding-decay 0 --amp 
--use-deterministic-algorithms --sync-bn 
--data-path ../../../imagenet_full/ILSVRC/Data/CLS-LOC/ 
--output-dir accvit_femto_imgnet_1k
\end{verbatim}

    \item ACC-ViT pico
\begin{verbatim}
CUBLAS_WORKSPACE_CONFIG=:16:8 torchrun --nproc-per-node=16 --nnodes=1 train.py 
--model accvit_p --epochs 400 --batch-size 160 --opt adamw --lr 1e-3 
--weight-decay 0.05 --lr-scheduler cosineannealinglr --lr-min 1e-5 
--lr-warmup-method linear --lr-warmup-epochs 32 --label-smoothing 0.1
--mixup-alpha 0.8 --cutmix-alpha 0.8 --clip-grad-norm 1.0 --interpolation bicubic
--auto-augment ta_wide --ra-magnitude 15 --ra-sampler --model-ema
--val-resize-size 224 --val-crop-size 224 --train-crop-size 224 
--model-ema-steps 32 --transformer-embedding-decay 0 --amp 
--use-deterministic-algorithms --sync-bn 
--data-path ../../../imagenet_full/ILSVRC/Data/CLS-LOC/ 
--output-dir accvit_pico_imgnet_1k
\end{verbatim}

\item ACC-ViT nano
\begin{verbatim}
CUBLAS_WORKSPACE_CONFIG=:16:8 torchrun --nproc-per-node=16 --nnodes=1 train.py 
--model accvit_n --epochs 400 --batch-size 128 --opt adamw --lr 1e-3 
--weight-decay 0.05 --lr-scheduler cosineannealinglr --lr-min 1e-5
--lr-warmup-method linear --lr-warmup-epochs 32 --label-smoothing 0.1
--mixup-alpha 0.8 --cutmix-alpha 0.8 --clip-grad-norm 1.0 --interpolation bicubic
--auto-augment ta_wide --ra-magnitude 15 --ra-sampler --model-ema
--val-resize-size 224 --val-crop-size 224 --train-crop-size 224 
--model-ema-steps 32 --transformer-embedding-decay 0 --amp 
--use-deterministic-algorithms --sync-bn
--data-path ../../../imagenet_full/ILSVRC/Data/CLS-LOC/ 
--output-dir accvit_nano_imgnet_1k
\end{verbatim}

\item ACC-ViT tiny
\begin{verbatim}
CUBLAS_WORKSPACE_CONFIG=:16:8 torchrun --nproc-per-node=16 --nnodes=1 train.py
--model accvit_t --epochs 400 --batch-size 64 --opt adamw --lr 1e-3
--weight-decay 0.05 --lr-scheduler cosineannealinglr --lr-min 1e-5
--lr-warmup-method linear --lr-warmup-epochs 32 --label-smoothing 0.1
--mixup-alpha 0.8 --cutmix-alpha 0.8 --clip-grad-norm 1.0 --interpolation bicubic
--auto-augment ta_wide --ra-magnitude 15 --ra-sampler --model-ema
--val-resize-size 224 --val-crop-size 224 --train-crop-size 224 
--model-ema-steps 32 --transformer-embedding-decay 0 --amp
--use-deterministic-algorithms --sync-bn
--data-path ../../../imagenet_full/ILSVRC/Data/CLS-LOC/
--output-dir accvit_tiny_imgnet_1k
\end{verbatim}

\clearpage

\end{itemize}

\end{strip}
\clearpage

\subsection{Finetuning on Medical Image Datasets}

In order to assess the transferability of the learned visual representations, we experimented on 3 medical image datasets of different modalities and sizes.

\begin{enumerate}
    \item \textbf{HAM10000} : The HAM10000 \cite{Tschandl2018TheLesions} dataset contains 10,015 dermatoscopic images from 7 different classes of pigmented skin lesions. The 7 classes are akiec (actinic keratoses and intraepithelial carcinomae), bcc (basal cell carcinoma), bkl (benign keratosis-like lesions), df (dermatofibroma), nv (melanocytic nevi), vasc (pyogenic granulomas and hemorrhage) and mel (melanoma). This dataset is quite imbalanced, nearly $2/3^{rd}$ of the images belong to the nv class, whereas akiec, df and vasc comprises $3.25\%$, $1.15\%$, and $1.4\%$, of the dataset respectively.

    \item \textbf{EyePACS} : The EyePACS dataset \cite{Cuadros2009EyePACS:Screening} is a source of high-resolution retina images taken under a variety of imaging conditions. In our experiments, we considered the 5,000 images of this dataset used for training in the Diabetic Retinopathy Detection competition, held at Kaggle (\url{https://www.kaggle.com/competitions/diabetic-retinopathy-detection/data}). The retinal images in this dataset are annotated into 5 grades of diabetic retinopathy, starting from none, spread to mild, moderate, severe, and up to proliferative level of diabetic retinopathy. This dataset is quite imbalanced as well, as almost $3/4^{th}$ of the retinal images have no diabetic retinopathy.

    \item \textbf{BUSI} : The BUSI dataset \cite{Al-Dhabyani2020DatasetImages} is a dataset of 780 breast ultrasound images from 600 patients. The images are categorized into 3 classes, namely, normal, benign, and malignant.
\end{enumerate}

We randomly partitioned the datasets into 60-20-20 train-validation-test splits. The models were fine-tuned using the training split and based on the validation loss the best checkpoints were obtained. Finally, the performance on the test split was assessed.

We adopted the finetuning strategy of MaxViT \cite{Tu2022MaxViT:Transformer} to finetune ACC-ViT and the baseline models on the medical image datasets. The learning rate was set to $5e^{-5}$ without any scheduler or warm-up. Other than RandAugment no other augmentation strategy, i.e., cutmix or mixup, was used. The values of label smoothing and gradient were set to 0.1 and 1.0, respectively. We finetuned the models for 100 epochs, longer than the usual 30 epochs. ACC-ViT managed to finetune rather quickly whereas MaxViT took longer on the medical images.

\clearpage

\section{Detailed Results}

\subsection{Finetuning on Medical Images}

Here, we present the detailed class-wise metrics and confusion matrices of the different models on the three different datasets.

From the overall results, a few points can be observed. 

\begin{enumerate}
    \item Across all the datasets, ACC-ViT predictions were consistently precise, which is evident from the highest precision scores.
    \item ACC-ViT has demonstrated a poor recall on the \textit{normal} class of the BUSI dataset, which has affected the overall score on that dataset. However, for the most important class of that dataset, i.e., \textit{malignent}, ACC-ViT actually managed to score the highest recall, $4.76\%$ higher than the best-performing ConvNeXt model.
    \item Notably, for the difficult \textit{mild} class of the EyePACS dataset ACC-ViT resulted in the highest recall, which is twice the second best method.
    \item MaxViT struggled quite a bit in predicting the rare classes. For example, it failed to classify any images of the df class from the HAM10000 dataset correctly.
\end{enumerate}

\begin{table*}[h]
\caption{Detailed results on the HAM10000 dataset}
\label{tbl:sham}
\footnotesize
\begin{tabular}{c|ccc|ccc|ccc|ccc|c}
\hline
               & \multicolumn{3}{c|}{ACC-ViT} & \multicolumn{3}{c|}{ConvNeXt} & \multicolumn{3}{c|}{Swin}   & \multicolumn{3}{c|}{MaxViT} &      \\
 &
  precision &
  recall &
  f1-score &
  precision &
  recall &
  f1-score &
  precision &
  recall &
  f1-score &
  precision &
  recall &
  f1-score &
  count \\ \hline
\textbf{akiec} & 0.8235  & 0.6462  & 0.7241  & 0.7069   & 0.6308  & 0.6667  & 0.7358  & 0.6000  & 0.6610 & 0.6857  & 0.3692  & 0.4800 & 65   \\
\textbf{bcc}   & 0.8785  & 0.9126  & 0.8952  & 0.8034   & 0.9126  & 0.8545  & 0.7851  & 0.9223  & 0.8482 & 0.6791  & 0.8835  & 0.7679 & 103  \\
\textbf{bkl}   & 0.8545  & 0.8545  & 0.8545  & 0.8586   & 0.7455  & 0.7981  & 0.8503  & 0.7227  & 0.7813 & 0.7026  & 0.7409  & 0.7212 & 220  \\
\textbf{df}    & 1.0000  & 0.7826  & 0.8780  & 0.8571   & 0.5217  & 0.6486  & 0.8667  & 0.5652  & 0.6842 & 0.0000  & 0.0000  & 0.0000 & 23   \\
\textbf{mel}   & 0.7860  & 0.8072  & 0.7965  & 0.7708   & 0.6637  & 0.7133  & 0.7031  & 0.6054  & 0.6506 & 0.6378  & 0.5291  & 0.5784 & 223  \\
\textbf{nv}    & 0.9586  & 0.9664  & 0.9625  & 0.9330   & 0.9761  & 0.9541  & 0.9260  & 0.9709  & 0.9479 & 0.9171  & 0.9567  & 0.9365 & 1341 \\
\textbf{vasc}  & 1.0000  & 0.9286  & 0.9630  & 0.9286   & 0.9286  & 0.9286  & 0.8966  & 0.9286  & 0.9123 & 0.8889  & 0.5714  & 0.6957 & 28   \\\hline
\begin{tabular}[c]{@{}c@{}}macro\\ avg\end{tabular} &
  0.9002 &
  0.8426 &
  0.8677 &
  0.8369 &
  0.7684 &
  0.7948 &
  0.8234 &
  0.7593 &
  0.7837 &
  0.6445 &
  0.5787 &
  0.5971 &
   \\
\begin{tabular}[c]{@{}c@{}}weighted\\ avg\end{tabular} &
  0.9205 &
  0.9206 &
  0.9200 &
  0.8918 &
  0.8957 &
  0.8918 &
  0.8784 &
  0.8832 &
  0.8786 &
  0.8318 &
  0.8462 &
  0.8354 &
   \\\hline
accuracy       & \multicolumn{3}{c|}{0.9206}  & \multicolumn{3}{c|}{0.8957}   & \multicolumn{3}{c|}{0.8832} & \multicolumn{3}{c|}{0.8462} &  \\ \hline   
\end{tabular}
\end{table*}

\begin{table*}[h]
\caption{Detailed results on the EyePACS dataset}
\label{tbl:sye}
\footnotesize
\begin{tabular}{c|ccc|ccc|ccc|ccc|c}
\hline
                       & \multicolumn{3}{c|}{ACC-ViT}   & \multicolumn{3}{c|}{ConvNeXt}  & \multicolumn{3}{c|}{Swin}      & \multicolumn{3}{c|}{MaxViT}    &       \\ \hline
                       & precision & recall & f1-score & precision & recall & f1-score & precision & recall & f1-score & precision & recall & f1-score & count \\ \hline
\textbf{none}         & 0.8970    & 0.9698 & 0.9320   & 0.8753    & 0.9766 & 0.9232   & 0.8626    & 0.9781 & 0.9167   & 0.8733    & 0.9804 & 0.9238   & 5162  \\
\textbf{mild}          & 0.4577    & 0.1881 & 0.2667   & 0.4839    & 0.0920 & 0.1546   & 0.5000    & 0.0348 & 0.0650   & 0.4894    & 0.0470 & 0.0858   & 489   \\
\textbf{moderate}      & 0.7269    & 0.6843 & 0.7050   & 0.7048    & 0.6295 & 0.6650   & 0.6812    & 0.6059 & 0.6413   & 0.6974    & 0.6144 & 0.6533   & 1058  \\
\textbf{severe}        & 0.5526    & 0.3600 & 0.4360   & 0.5271    & 0.3886 & 0.4474   & 0.5046    & 0.3143 & 0.3873   & 0.4929    & 0.3943 & 0.4381   & 175   \\
\textbf{proliferative} & 0.6866    & 0.6479 & 0.6667   & 0.7500    & 0.5282 & 0.6198   & 0.8202    & 0.5141 & 0.6320   & 0.6786    & 0.5352 & 0.5984   & 142   \\ \hline
\begin{tabular}[c]{@{}c@{}}macro\\ avg\end{tabular}    & 0.6642 & 0.5700 & 0.6012 & 0.6682 & 0.5230 & 0.5620 & 0.6737 & 0.4894 & 0.5285 & 0.6463 & 0.5143 & 0.5399 &  \\
\begin{tabular}[c]{@{}c@{}}weighted\\ avg\end{tabular} & 0.8280 & 0.8507 & 0.8338 & 0.8112 & 0.8390 & 0.8128 & 0.8003 & 0.8305 & 0.7971 & 0.8067 & 0.8367 & 0.8061 &  \\ \hline
accuracy               & \multicolumn{3}{c|}{0.8507}    & \multicolumn{3}{c|}{0.8390}    & \multicolumn{3}{c|}{0.8305}    & \multicolumn{3}{c|}{0.8367}    &   \\ \hline   
\end{tabular}
\end{table*}

\begin{table*}[h]
\caption{Detailed results on the BUSI dataset}
\label{tbl:sbusi}
\footnotesize
\begin{tabular}{c|ccc|ccc|ccc|ccc|c}
\hline
                   & \multicolumn{3}{c|}{ACC-ViT}   & \multicolumn{3}{c|}{ConvNeXt}  & \multicolumn{3}{c|}{Swin}      & \multicolumn{3}{c|}{MaxViT}    &       \\ \hline
                   & precision & recall & f1-score & precision & recall & f1-score & precision & recall & f1-score & precision & recall & f1-score & count \\\hline
\textbf{benign}    & 0.8261    & 0.8736 & 0.8492   & 0.8539    & 0.8736 & 0.8636   & 0.8161    & 0.8161 & 0.8161   & 0.7358    & 0.8966 & 0.8083   & 87    \\
\textbf{malignent} & 0.7674    & 0.7857 & 0.7765   & 0.8611    & 0.7381 & 0.7949   & 0.8333    & 0.7143 & 0.7692   & 0.7273    & 0.5714 & 0.6400   & 42    \\
\textbf{normal}    & 0.8095    & 0.6296 & 0.7083   & 0.6774    & 0.7778 & 0.7241   & 0.6061    & 0.7407 & 0.6667   & 0.7059    & 0.4444 & 0.5455   & 27    \\\hline
\begin{tabular}[c]{@{}c@{}}macro\\ avg\end{tabular}    & 0.8010 & 0.7630 & 0.7780 & 0.7975 & 0.7965 & 0.7942 & 0.7518 & 0.7570 & 0.7507 & 0.7230 & 0.6375 & 0.6646 &  \\
\begin{tabular}[c]{@{}c@{}}weighted\\ avg\end{tabular} & 0.8074 & 0.8077 & 0.8052 & 0.8253 & 0.8205 & 0.8210 & 0.7844 & 0.7756 & 0.7776 & 0.7284 & 0.7308 & 0.7175 &  \\\hline
accuracy           & \multicolumn{3}{c|}{0.8077}    & \multicolumn{3}{c|}{0.8205}    & \multicolumn{3}{c|}{0.7756}    & \multicolumn{3}{c|}{0.7308}    &  \\ \hline    
\end{tabular}
\end{table*}

\begin{figure*}[h]
     \centering
     \begin{subfigure}[b]{0.99\textwidth}
         \centering
         \includegraphics[width=\textwidth]{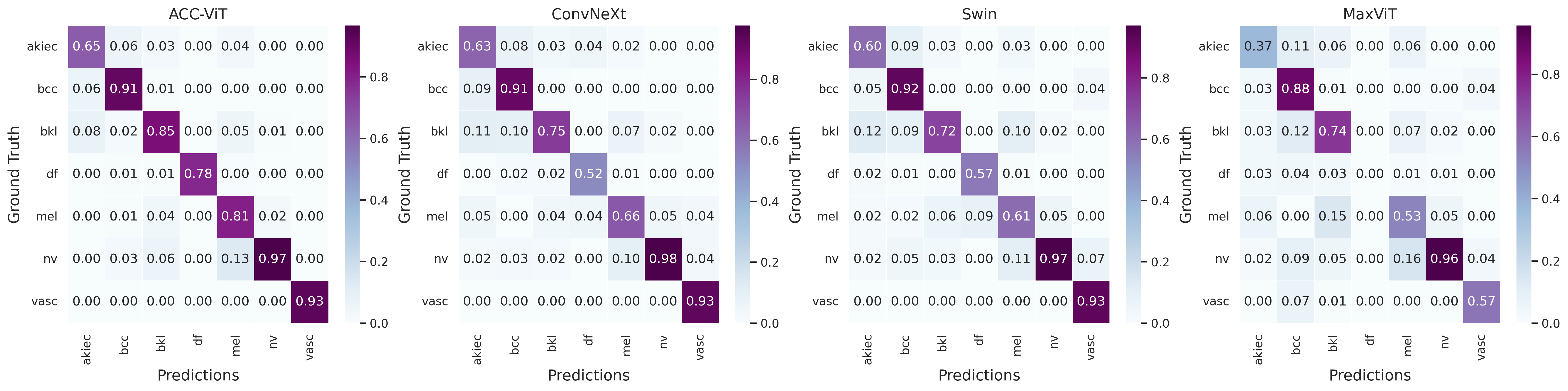}
         \caption{HAM10000 dataset}
     \end{subfigure}
     \hfill
     \begin{subfigure}[b]{0.99\textwidth}
         \centering
         \includegraphics[width=\textwidth]{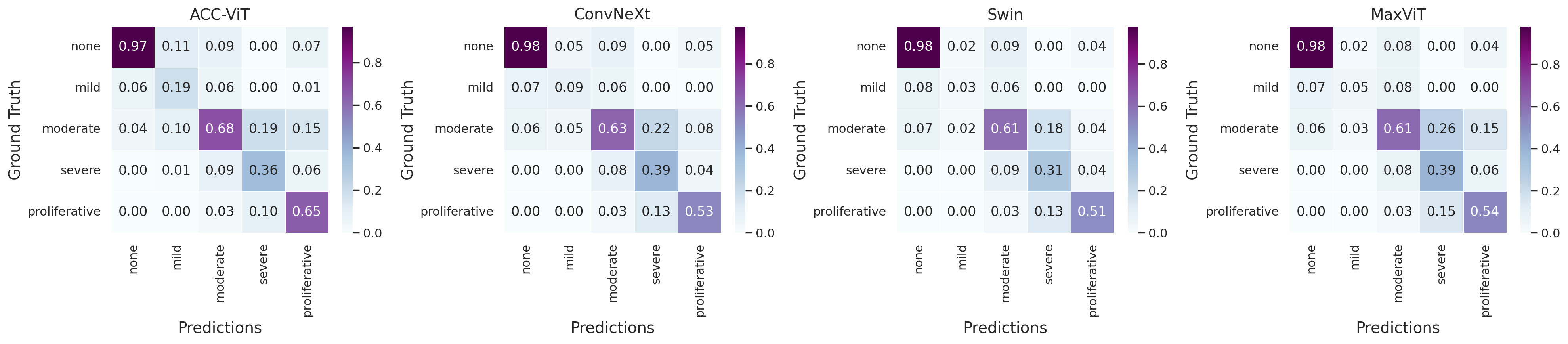}
         \caption{EyePACS dataset}
     \end{subfigure}
     \hfill
     \begin{subfigure}[b]{0.7\textwidth}
         \centering
         \includegraphics[width=\textwidth]{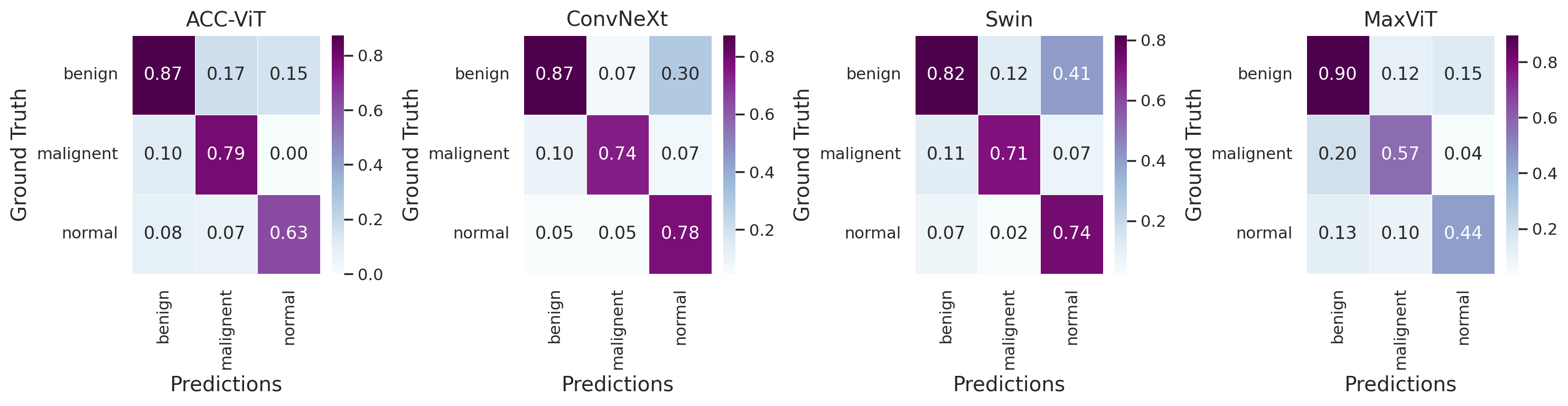}
         \caption{BUSI dataset}         
     \end{subfigure}
        \caption{Confusion Matrix of the different models on the three datasets.}
        \label{fig:three graphs}
\end{figure*}

\begin{table*}[b]
\centering
\caption{Ablation Study}
\label{tbl:ablation}
\begin{tabular}{lccc}
\hline
      & \#params(M) & FLOPs(G) & Acc \\ \hline
Atrous Attention with MBConv&   16.907          &   4.862        & 79.498    \\ \hline
Inroducing Atrous Convolution &         19.33    &  4.939         & 80.017    \\ \hline
Shared MLP across attentions &         14.585      & 3.473     & 81.523     \\ \hline
 Adaptive Gating in Parallel branches & 16.649 & 3.812 & 82.412 \\ \hline
Replacing MLP with ConvNext & 17.025 & 3.844 & 81.850 \\ \hline
\end{tabular}
\end{table*}

\section{Ablation Study}
\label{sec:abla}

Table \ref{tbl:ablation} presents an ablation study focusing on the contributions of the different design choices in our development roadmap. We conducted our design discovery and ablation study on the nano variant, i.e., $\sim 17$ M parameter model.

We started with Atrous Attention in conjunction with MBConv, which resulted in $79.5 \%$ accuracy. Later, we used parallel atrous convolutions, increasing the accuracy only by $0.5\%$ with a considerable increase in parameters. Sharing the MLP layer across attentions, turned out quite beneficial, substantially reducing the FLOPs and increasing the accuracy. Up to this moment, the model, merely averaging the different regional information, would apparently learn the parameters quickly and reach a plateau. Implementing the gating function improved this scenario, as the model learned to focus on the key regional information dynamically, based on the input. We also attempted to replace the MLP layer with ConvNext layer, similar to MOAT, but that did not yield a satisfactory outcome and thus was discarded. 

\clearpage

\end{document}